\documentclass[11pt,a4paper]{article}
\usepackage[hyperref]{naaclhlt2019}
\usepackage{times}
\usepackage{CJKutf8}
\usepackage{latexsym}
\usepackage{url}
\usepackage[multiple]{footmisc}
\usepackage{graphicx}
\usepackage{subfigure}
\usepackage{changepage}
\usepackage{mathdots}
\usepackage{comment}
\usepackage{amsmath}
\usepackage{enumitem}

\aclfinalcopy 

\title{On conducting better validation studies of automatic metrics in natural language generation evaluation}

\author{Johnny Tian-Zheng Wei \\
  University of Massachusetts Amherst \\ College of Natural Sciences \\
  {\tt jwei@umass.edu}
\\}

\date{}

\begin{document}
\maketitle
\begin{abstract}
  Natural language generation (NLG) has received increasing attention, which has highlighted evaluation as a central methodological concern. Since human evaluations for these systems are costly, automatic metrics have broad appeal in NLG. Research in language generation often finds situations where it is appropriate to apply existing metrics or propose new ones. The application of these metrics are entirely dependent on validation studies - studies that determine a metric's correlation to human judgment. However, there are many details and considerations in conducting strong validation studies. This document is intended for those validating existing metrics or proposing new ones in the broad context of NLG: we 1) begin with a write-up of best practices in validation studies, 2) outline how to adopt these practices, 3) conduct analyses in the WMT'17 metrics shared task\footnote{Our jupyter notebook containing the analyses is available at \url{https://github.com}}, and 4) highlight promising approaches to NLG metrics 5) conclude with our opinions on the future of this area.
\end{abstract}

\section{Introduction}

Increasing interest in tasks that require generating natural language such as image captioning \cite{DBLP:conf/eccv/LinMBHPRDZ14}, dialogue \cite{DBLP:journals/corr/VinyalsL15}, and style transfer \cite{DBLP:conf/aaai/FuTPZY18} highlight evaluation as a central methodological concern. With human involvement, a system can be evaluated extrinsically \citep[how well does the system fulfill its intended purpose? e.g.][]{DBLP:journals/ai/ReiterRO03} or intrinsically (what is the quality of the output?) In domains such as machine translation (MT), a system's extrinsic value is both hard to define and measure, and intrinsic human judgments of a system's output quality have been the main indicator of progress in the field. \cite{bojar-etal_Cracker:2016}

This paper focuses on those domains best evaluated intrinsically. Since acquiring intrinsic human judgments is costly, automatic metrics, which both are computed automatically and correlate highly with human judgment, are ideal. If sufficiently correlated, a metric can be used as a surrogate evaluation, which may be useful in developmental cycles. Therefore, the application of such metrics are dependent on studies of their validity \citep[][how well does the metric correlate with human judgment?]{DBLP:journals/coling/Reiter18}. In MT, BLEU \cite{DBLP:conf/acl/PapineniRWZ02} has seen widespread use, and, consequently, its validity has been extensively studied. \cite[][inter alia]{ DBLP:conf/eacl/Callison-BurchOK06}

\begin{figure}[t!]
\begin{center}
\begin{adjustwidth}{-0.4cm}{}
\begin{tabular}{|p{1.7cm}|p{1.7cm}|p{1.7cm}|p{0.7cm}|p{0.7cm}|p{0.7cm}|}
\hline
Source & Reference & Output & DA & \hspace*{-0.55em} BEER \\
\hline
\hline
\tiny\begin{CJK*}{UTF8}{gbsn}双方很难，甚至不可能重新建立真正的信任。\end{CJK*} & \tiny rebuilding real trust will be hard , perhaps impossible . & \tiny it is difficult , if not impossible , to re-establish real trust . & \small 0.55 & \small 0.39 \\
\hline
\tiny\begin{CJK*}{UTF8}{gbsn}如何在持枪攻击中使用马伽术保护自己\end{CJK*} & \tiny how to defend yourself 
from gun attacks using krav ... & \tiny how to use marcella to protect himself in a gun attack & \small -0.85 & \small 0.42 \\
\end{tabular}
\end{adjustwidth}
$\vdots$
\end{center}
\vspace{-1.0em}
\caption{\label{exs} Examples from the WMT'17 metrics task for \texttt{\small zh-en} translation evaluation. Outputs produced from \citet{DBLP:conf/wmt/SennrichBCGHHBW17}. The DA score is a mean human judgment of translation quality. More on DA (Direct Assessment) in \S 2.2. Scores of a participating metric \citep[BEER;][]{DBLP:conf/wmt/StanojevicS14} are shown. Metrics aim to achieve high correlation with DA scores. More on task details in \S 2.}
\vspace{-1.2em}
\end{figure}

These automatic metrics are generally appealing to natural language generation (NLG) domains. During 2005-2014, \citet{DBLP:conf/enlg/GkatziaM15} found that a significant portion of NLG research in ACL reported results from automatic metrics. Research in under-explored NLG domains have begun with proposals of both models and novel evaluation metrics. \cite{DBLP:conf/aaai/FuTPZY18, DBLP:journals/corr/abs-1811-05701} For the application of any existing metric (e.g. BLEU) or newly proposed metric to bear validity in various domains, researchers have attempted to conduct validation studies, for tasks such as surface realization \cite{DBLP:conf/emnlp/NovikovaDCR17}, open-domain dialogue \cite{DBLP:conf/emnlp/LiuLSNCP16}, and image captioning \cite{DBLP:conf/eacl/KilickayaEIE17}, at times reporting negative results.

However, there are many considerations when designing a validation study. There are at least two aspects that experimenters should be cognizant about in conducting a robust validation study: first are the assumptions made; the second is statistical methodology. Not unlike testing of our models, validation of our metrics should also be approached with rigor. At the time of writing, the metrics shared task in the Conference of Machine Translation \citep[WMT;][]{DBLP:conf/wmt/BojarFFGHKM18} annually conducts strong validation studies and we recommend adopting best practices from this domain.

\begin{figure}
    \centering
    \includegraphics[scale=0.1]{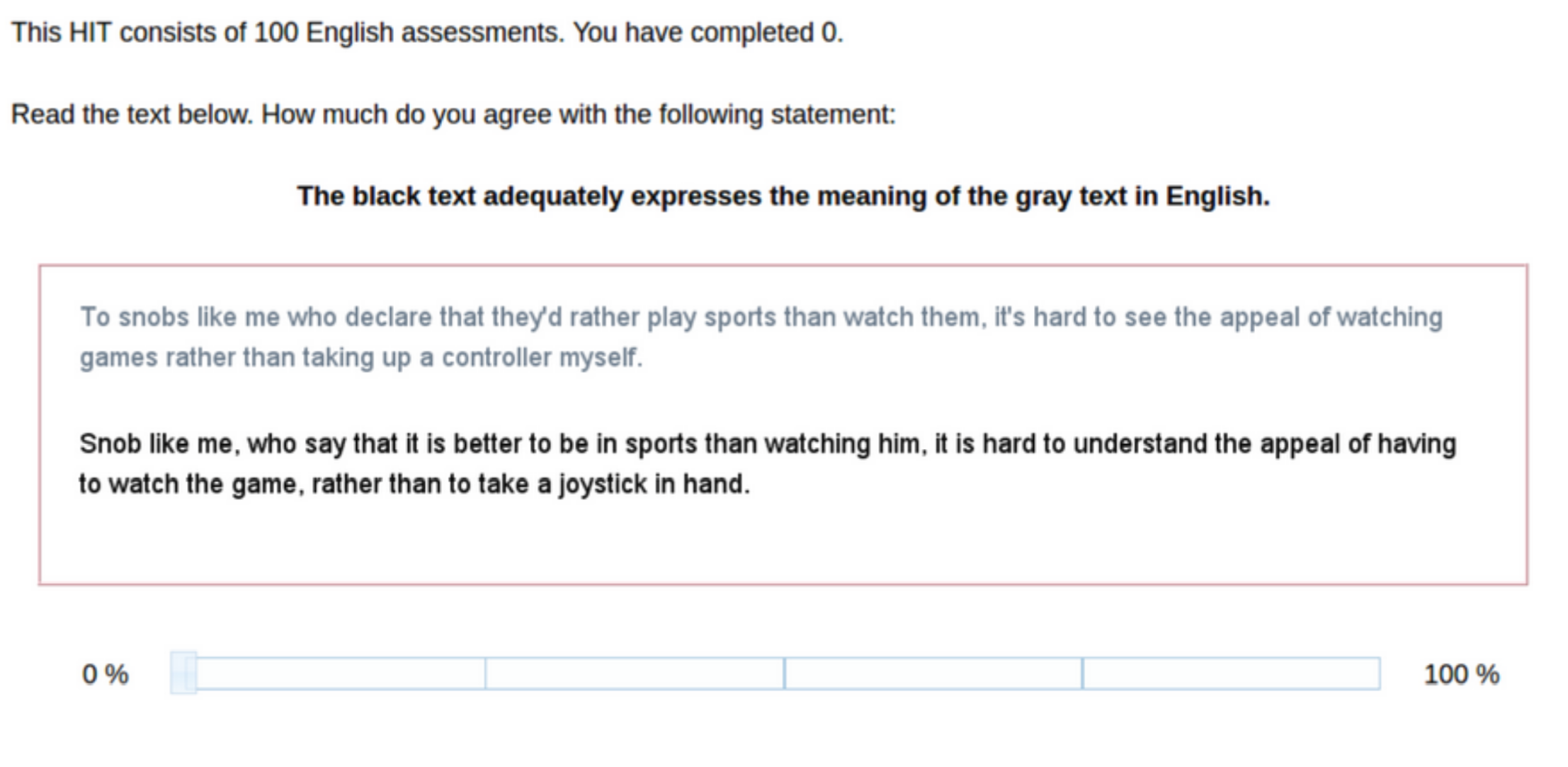}
    \caption{\label{da-hit} The interface for collection of {\it direct assessment} (DA) scores from Amazon Mechanical Turk workers. Figure taken from \citet{DBLP:conf/wmt/BojarCFGHHHKLLM17}. Each worker rates system outputs on a continuous 0-100 scale, which are converted to deviations from the worker's mean to normalize over scoring strategies.  }
    \label{fig:my_label}
\end{figure}


This document is intended for those validating existing metrics or proposing new ones in NLG. To this end, we present: 
\begin{itemize}[topsep=0.3em]
    \itemsep0em 
    \item an overview of the WMT metrics shared task validation procedures (\S 2) and an outline on how to adopt these best practices, generally, to other NLG domains (\S 3). 
    
    \item proposals of basic analyses of metrics scoring, complemented by analysis on existing data from the WMT'17 metrics task (\S 4). 
    
    \item a literature review in both metrics (\S 5) and their analysis, (\S 6) concluding with the authors' opinions on directions of the area (\S 7).
\end{itemize}

\section{WMT metrics shared task}

The metrics shared task \cite{DBLP:conf/wmt/MaBG18} utilizes the WMT evaluation results of the popular translation shared task. The data provided from the translation shared task is diverse, including output from a range of state-of-the-art systems. Over the years, WMT also has developed robust statistical methodology for collecting human judgments and significance testing of metric performance. For these reasons, the results from the metrics shared task are particularly strong. We highlight relevant methodological aspects in the following sections.

\begin{figure}
    \centering
    \includegraphics[scale=0.16]{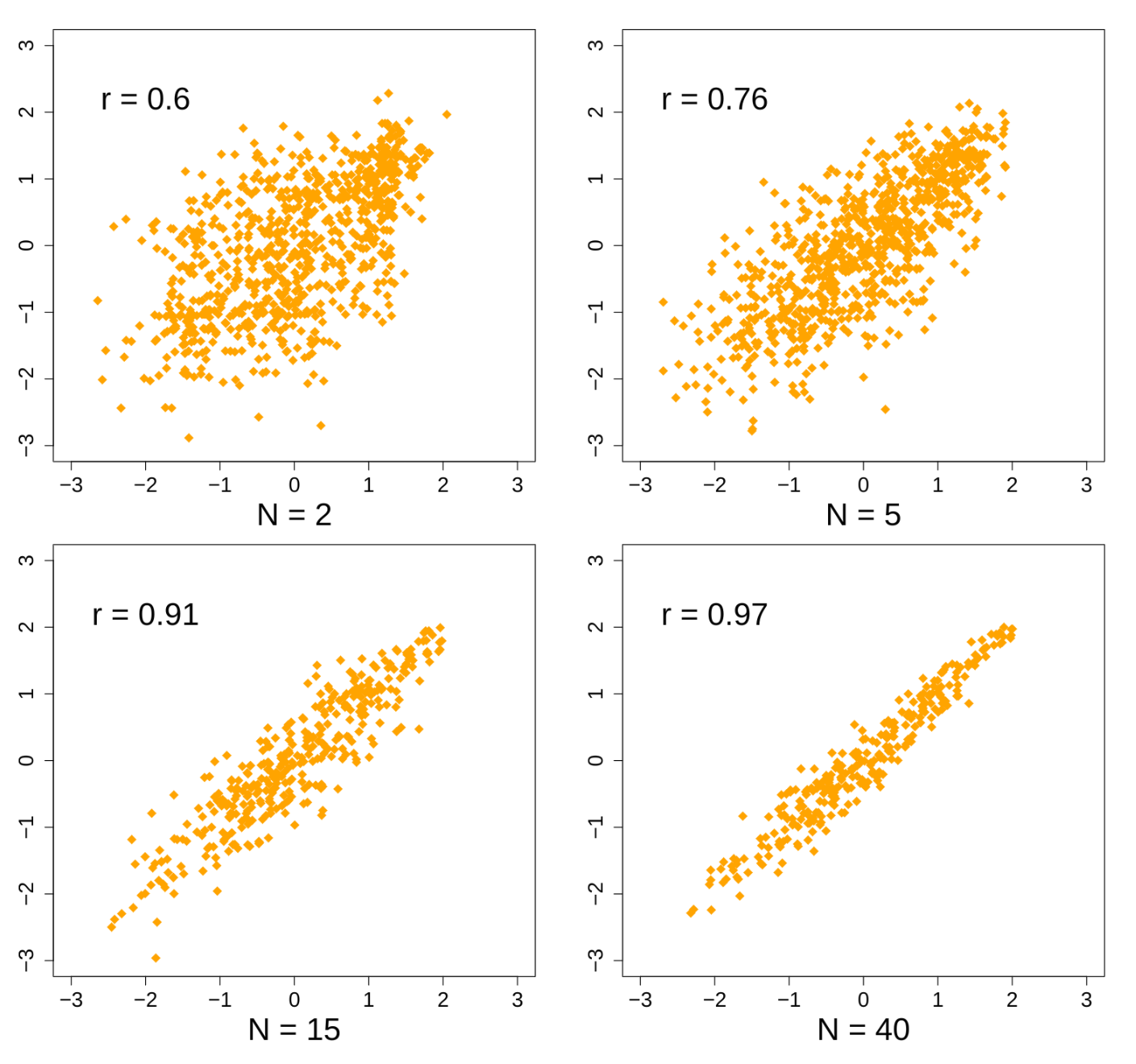}
    \caption{\label{da-num} The effect of the number of assessors on the consistency of DA scores. Figure taken from \citet{DBLP:conf/naacl/GrahamBM15}. The $x$-axis is a sample mean of translation quality calculated from $n$ judgments, and the $y$-axis is a true mean estimated from a much larger population.}
\end{figure}

\subsection{Direct Assessment}

Beginning in WMT'17, the WMT human evaluation campaign has adopted {\it direct assessment} \cite[DA;][]{DBLP:conf/naacl/GrahamBM15} as the primary human evaluation. \cite{DBLP:conf/wmt/BojarCFGHHHKLLM17} DA scores are formulated on the law of large numbers - a sample mean of many human judgments is close to the population mean, which represents an intrinsic property of the translation quality.

\begin{figure*}
\centering
\begin{subfigure}{}
\includegraphics[scale=1]{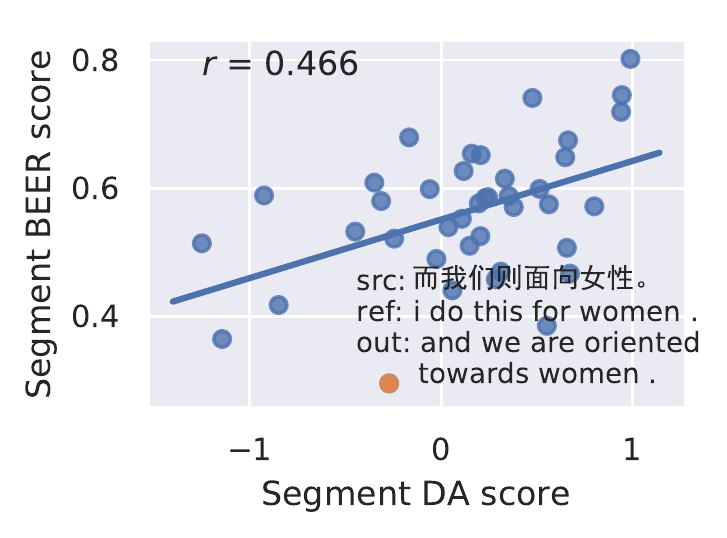}
\end{subfigure}
~
\begin{subfigure}{}
\includegraphics[scale=1]{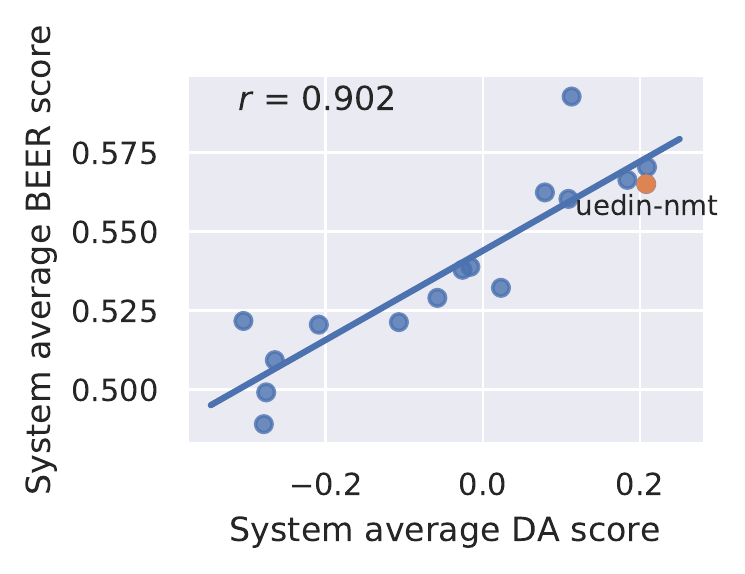}
\end{subfigure}
\caption{\label{sys-seg-corr}  Sample Pearson $r$ of segment/system-level BEER and DA scores from the WMT'17 metrics task for \texttt{\small zh-en} translation evaluation. {\bf Left.} Each point denotes a test example from the metrics task. Examples are a subset from the \texttt{uedin-nmt} system \cite{DBLP:conf/wmt/SennrichBCGHHBW17}. {\bf Right.} Each point denotes a system participating in the \texttt{\small zh-en} translation task. DA and BEER scores have been averaged over all outputs for a given system.}
\end{figure*}

However, we do not know the variance of human scores for a given translation, so an appropriate sample size $n$ cannot be calculated. In addition, the variance in judgments can vary considerably across translations. To overcome this, DA uses a two step process: 1.  Empirically determine the number of assessors $n$ needed per translation for the desired consistency. 2. Collect $n$ human judgments for remaining system outputs. 

Refer to Figure \ref{da-hit}. DA uses a continuous sliding bar from 1-100, which are averaged over many workers for a sample mean. \cite{DBLP:conf/acllaw/GrahamBMZ13} Refer to Figure \ref{da-num}. In the first phase of DA, a large ($n=100$) number of assessments are collected for a small number of output translations. Compute a simulated DA score for each segment with the first $i$ judgments, and estimate a true mean with the remaining $n-i$ judgments. For the $i$ where the correlation between DA scores and ``true mean'' is sufficient, gather $i$ judgments for each segment in the full data collection.

Several aspects of DA have made it appealing for the WMT shared tasks. These points are stated in \citet{DBLP:conf/wmt/BojarCFGHHHKLLM17}: 1) DA score are more consistent than relative rankings of system output, which are often contradictory 2) The samples means from DA scores are absolute, facilitating comparisons across translations 3) Sliding bar judgments can be collected from crowdsourcing websites, and there are effective measures for quality control. 

\subsection{System and sentence-level correlation}

Direct Assessment gives human judgment scores on a sentence (or segment) level. System level quality is then defined as the average of DA scores of a system's output over an entire test set, and is how the rankings for the annual WMT translation task are produced. (\citeauthor{DBLP:conf/wmt/BojarFFGHKM18}, \citeyear{DBLP:conf/wmt/BojarFFGHKM18, DBLP:conf/wmt/BojarCFGHHHKLLM17}) Refer to Figure \ref{sys-seg-corr}. Therefore, a metric can be evaluated on two correlations: at the segment-level - between segments' metric scores and DA scores, or system-level - between systems' average DA scores and aggregate metric scores (most commonly a mean).

There are several notable distinctions between the two. For a metric to correlate highly on the system level does not imply it correlates on the segment level. In fact, in the MT domain, baselines such as BLEU have high correlations on the system-level for \texttt{*-en} translation evaluation ($r > 0.9$), but have low segment-level correlations \cite[$r < 0.5$,][]{DBLP:conf/wmt/MaBG18}. This is also seen for the SPICE \cite{DBLP:conf/eccv/AndersonFJG16} metric in image captioning. Intuitively, a metric may only be competent in penalizing bad output, but cannot differentiate between average and good output (Refer to \S 4.1 for an analysis of BLEU). This causes the discrepancy in low correlation at the segment-level but high at the system-level. \cite{DBLP:conf/emnlp/NovikovaDCR17,DBLP:conf/acl/LiangCM18}


In practice, system-level correlation is relevant. Research will often report results of system-level BLEU scores to show the effectiveness of a new system over baselines or existing literature. \cite{DBLP:conf/emnlp/Koehn04} When making decisions about hyperparameters or model architectures, we rely on metric-produced system rankings. \cite{D17-1151} However, existence of metrics with high segment-level correlations opens up research questions e.g. can we use such a metric as an alternative training objective? \cite{DBLP:journals/corr/RanzatoCAZ15}

\begin{figure}
    \centering
    \includegraphics[scale=0.17]{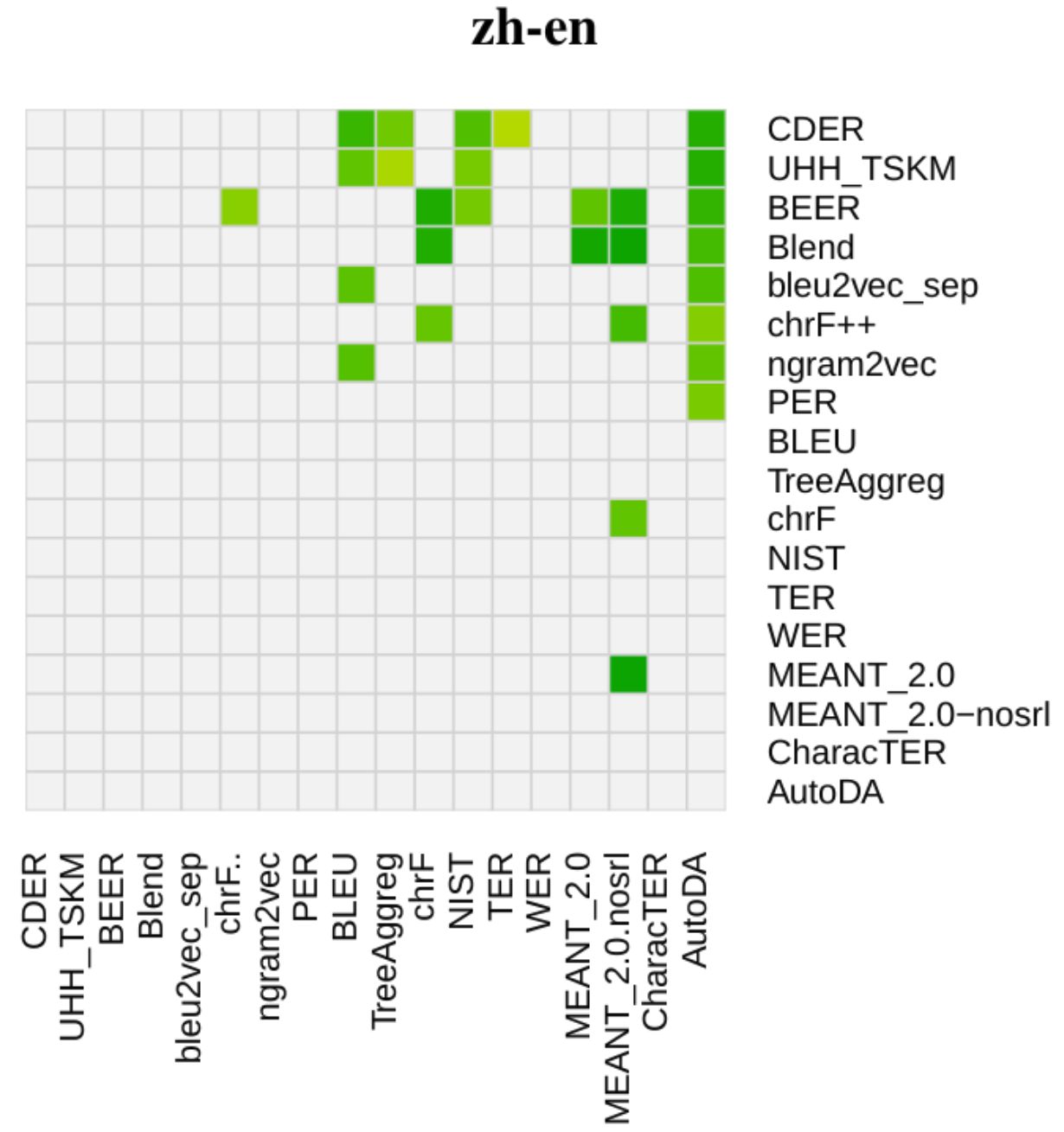}
    \caption{\label{williams} $p$-values from the William's test on all pairs of metrics participating in system-level, \texttt{zh-en} translation evaluation in WMT'17. A green cell denotes ($p < 0.05$). Figure taken from \cite{DBLP:conf/wmt/BojarGK17}. Significance in cell $(i,j)$ denotes metric in row $i$ significantly outperformed metric in column $j$.  }
    \label{fig:my_label}
\end{figure}

\subsection{Pearson correlation}

The emerging consensus in WMT is the use of Pearson correlation\footnote{\url{https://libguides.library.kent.edu/SPSS/PearsonCorr}} in segment and system-level evaluation of metrics. Given $n$ paired data points $\{(x_1, y_1), ..., (x_n, y_n)\}$, the sample Pearson correlation is defined as:
$$r_{xy} = \frac{\sum^n_{i=1}(x_i-\overline{x})(y_i-\overline{y})}{\sqrt{\sum^n_{i=1}(x_i-\overline{x})^2}\sqrt{\sum^n_{i=1}(y_i-\overline{y})^2}}$$
where $\overline{x}$ and $\overline{y}$ are the samples means for $x_i$ and $y_i$, respectively. This correlation measures a linear association between two ordinal variables and ranges from $[-1, 1]$. In segment-level metric validation, $(x_i, y_i)$ may represent the DA and metric scores of a system translation. On the system level, $(x_i, y_i)$ may represent the average DA and aggregated metric scores of an MT system.

There are at least three reasons for motivation of the use of Pearson coefficient over Kendall's and Spearman. (These reasons are stated across \citeauthor{DBLP:conf/naacl/GrahamBM15} \citeyear{DBLP:conf/naacl/GrahamBM15, DBLP:conf/emnlp/GrahamB14}) 1) The linear measurement of Pearson correlation measure is more sensitive than rank-based correlations (i.e. Spearman), which measure any monotonic relationship. 2) Pearson correlations are absolute, facilitating comparisons across testing settings. 3) Significance testing of differences in Pearson values are possible (see \S 2.4).

\subsection{Significance tests for correlations}

For the sample Pearson correlation $r$ calculated between metric and DA scores, there are two significance tests that are often applied. The first test (which is \emph{not} used in the WMT metrics task) calculates a $p$-value to reject the null hypothesis $H_0: r=0$. An unlikely null hypothesis means your metric has some true non-zero correlation. This is a common test, and $p$-values will be provided by most standard statistical packages.\footnote{\texttt{cor.test} in \textsf{R} or \texttt{stats.pearsonr} in \texttt{scipy}.}

The WMT metrics task uses the William's test \cite{williams1959regression}, which tests for the significance of differences in Pearson correlation. Concretely, a $p$-value is calculated to reject $H_0:Corr(X_1, X_3) = Corr(X_2, X_3)$, where $X_1$ and $X_2$ are correlated. In our case, $X_1$ and $X_2$ are generalizations of two different metric scores, and $X_3$ of DA scores. \cite{DBLP:conf/naacl/GrahamBM15} An unlikely null hypothesis means that it is likely one metric is better than the other.

Refer to Figure \ref{williams}. In the WMT metrics task, the winners are declared as those metrics which were not significantly outperformed by any other metric. In WMT'17 for \texttt{zh-en}, 9 different metrics were co-winners. \cite[][count the empty columns in Figure \ref{williams}]{DBLP:conf/wmt/BojarGK17}




\section{Conducting a validation study}

We assume that a suitable NLG task has been chosen for evaluation. The section will outline considerations and statistical methodology step-by-step in your validation study. We synthesize from existing literature we read, and borrow heavily from the design of the WMT metrics shared task. 

\subsection{Collecting diverse system output}

The results of your validation study will vary greatly based on testing conditions of the metrics. Conclusions generalize best to similar conditions in practice, and so it is important to cover as much ground as possible.

\begin{itemize}[topsep=0.3em]
    \itemsep0em 
    \item If you are using Pearson correlation (recommended, refer to \S 2.2) to measure a metric's system-level correlation, choose at least five (5) systems. Five points is the least amount of data points to make a statistically significant conclusion ($p < 0.05$) that a correlation is non-zero.\footnote{\url{https://ehudreiter.com/2018/07/10/how-to-validate-metrics/}} \cite{DBLP:journals/coling/Reiter18}
    
    \item Produce outputs from a mix of baselines and state-of-the art systems using a variety of approaches. In MT, a few popular approaches may include transfer-based, statistical, and neural systems. It is known, for instance, that BLEU correlates poorly with rule-based systems, \cite{DBLP:conf/eacl/Callison-BurchOK06} and the exclusion of these systems have caused correlations between BLEU and human judgments to increase. \cite{DBLP:journals/coling/Reiter18}
    
    \item If possible, include a few synthetic variations of a system, ideally variations seen in practice. \cite{DBLP:conf/eacl/Callison-BurchOK06} One such variation could be several identical models with ablated subsets of the training data.
    
    \item Report characteristics of the test set system output is elicited from. Metrics using linguistic resources (e.g. WordNet, parsers, taggers) will be sensitive to the language they are applied to. \cite{DBLP:conf/eacl/KilickayaEIE17}
\end{itemize}

\subsection{Collecting consistent human judgments}

Consistent human judgments are necessary to accurately evaluate metrics. Collect data in a manner that is replicable and publicly release the data.

\begin{itemize}[topsep=0.3em]
    \itemsep0em 
    \item Choose a question to elicit intrinsic quality judgment from humans. Unfortunately, the choice of question is nearly art and considerations may be philosophical. \cite{DBLP:journals/jair/GattK18}. However, a question that elicits consistent judgment saves effort in annotation, and may mean the question reflects a true intrinsic property. Note that BLEU was originally validated against human judgment of ``general translation quality.'' \cite{DBLP:conf/acl/PapineniRWZ02}
    
    \item Choose one intrinsic quality question. In WMT, aspects of adequacy and fluency were previously judged separately but later abandoned. \cite{DBLP:conf/wmt/BojarCFGHHJKLMN16} This is not ideal because two forms of results are confusing, and if you expect a metric to correlate with both aspects, only one question is needed.
    
    \item Direct assessment (\S 2.1) collects consistent human judgments, and can be crowdsourced relatively hassle-free through Amazon Mechanical Turk. \cite{DBLP:conf/naacl/GrahamBM15, DBLP:conf/wmt/BojarCFGHHHKLLM17}. This collection methodology has also been successfully applied NLG domains outside of MT. Refer to \citet{10.1371/journal.pone.0202789} for an example in video captioning.
\end{itemize}

\subsection{Producing automatic metric scores}

\begin{itemize}[topsep=0.3em]
    \itemsep0em
    \item Use consistent tokenization across all metrics. Most metrics, especially $n$-gram based metrics, will be affected by tokenization. \cite{DBLP:conf/wmt/Post18}
    
    \item You will likely be using BLEU and/or sentBLEU as a baseline. If so, note factors affecting scores (e.g. preprocessing, $n$-gram weighting, length penalty), and that S{\small ACRE}BLEU is an existing tool to manage these parameters. \cite{DBLP:conf/wmt/Post18}
\end{itemize}

\subsection{Conducting significance tests}

Applying robust statistical methodology will allow sound conclusion to be drawn about the performance of our metrics.

\begin{itemize}[topsep=0.3em]
    \itemsep0em 
    \item Don't leave anything to chance - use the William's test (\S 2.4) to test for significance in increase of Pearson correlation. \citet{DBLP:conf/emnlp/GrahamB14} provides an open source implementation at \url{https://github.com/ygraham/significance-williams}.
    
    \item Report both system and segment-level results with significant tests. (\S 2.2) You may also want to use Kendall's $\tau$ or Spearman's $\rho$ as secondary metric evaluations. (\S 4)
\end{itemize}

\section{Analyses of metrics in WMT'17}


This section demonstrates several basic analysis methods of metric scoring on both the segment and system level. We will exclusively perform these analyses on BLEU and its sentence-level variant sentence-BLEU\footnote{\url{https://github.com/moses-smt/mosesdecoder/blob/master/mert/sentence-bleu.cpp}}, \cite{Koehn:2007:MOS:1557769.1557821} for demonstration and for insight in a widely applied baseline metric.

\begin{figure}[t]
    \centering
    \includegraphics[scale=1]{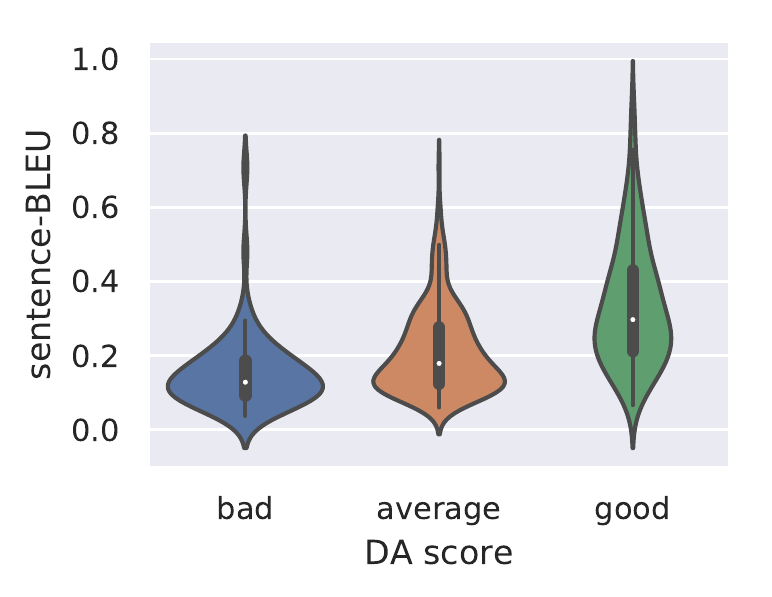}
    \caption{\label{bleu-cond-dist} Distribution of sentence-BLEU scores conditional on bin of DA score for \texttt{\small zh-en} translation evaluation in WMT'17. The three intervals for binning are $\small(-1.73, -0.749]<(-0.749, 0.228]<(0.228, 1.205]$. Every bin has a near equal number of points. }
\end{figure}

\begin{table*}[t!]
    \centering
    \begin{tabular}{|c|p{3.6cm}|p{3.6cm}|p{3.6cm}|p{1.0cm}|p{1.0cm}|}
        \hline
        Bin & Source & Reference & Output & DA & \small sentence-BLEU  \\
        \hline
        \hline
        ~ & \tiny
        \begin{CJK*}{UTF8}{gbsn}如果两家工厂关闭，则电力市场的需求量会大大减少。\end{CJK*} & \tiny if both of those plants go from the market that 's a significant reduction in demand in the [ electricity ] market . & \tiny if the two factories closed , the power market demand will be greatly reduced . & 0.355 & 0.066 \\
        \cline{2-6}
        good & \tiny \begin{CJK*}{UTF8}{gbsn}上次产油国召开会议已经是4月份的事情，OPEC成员国未能就任何措施达成协议。\end{CJK*} & \tiny it was april when the oil-producing countries had meeting . no agreement was reached among the opec member ... & \tiny the last time oil producers are meeting in april , opec member states that have failed to agree on any measures to reach an ...	 & 0.247 & 0.079 \\
        \cline{2-6}
        ~ & \tiny \begin{CJK*}{UTF8}{gbsn}小鹏和吴言在一个大群里相识，因为两人都是积极发言者，很快就熟络起来了。\end{CJK*} & \tiny xiaopeng and wuyan meet each other in the group chatting . they two get familiar with each other because they are active ... & \tiny xiao peng and wu met in a large group , because both were active speakers and soon became familiar . & 0.537 & 0.087 \\
        \hline
        \hline
        ~ & \tiny \begin{CJK*}{UTF8}{gbsn}朗兹曼写道。\end{CJK*} & \tiny lanzmann wrote . & \tiny \begin{CJK*}{UTF8}{gbsn}朗兹曼\end{CJK*} wrote . & -1.245 & 0.707 \\
        \cline{2-6}
        bad & \tiny \begin{CJK*}{UTF8}{gbsn}云南省首届青运会13日开赛开幕式打民族牌展示青春活力\end{CJK*} & \tiny the opening ceremony of the first youth games of yunnan province on the 13th day of the month showed youth vitality & \tiny the opening ceremony of the opening ceremony of the first olympic games of yunnan province on the opening ceremony of yunnan province & -1.081 & 0.481 \\
        \cline{2-6}
        ~ & \tiny \begin{CJK*}{UTF8}{gbsn}远在千里之外，巨嘴鸟格雷西亚的故事感动了著名纪录片导演葆拉·埃雷迪亚和探索新闻频道制片人约翰·霍夫曼。\end{CJK*} & \tiny thousands of miles away , the story of the toucan gracia touched the famous documentary director paula el reidia and the news channel producer john hoffman . & \tiny far more than thousands of thousands , the story of greecia is moved by the famous documentary director paula mareda and the exploration of the producer of news ... & -0.776 & 0.282 \\
        \hline
    \end{tabular}
    \caption{Selections of lowest scoring BLEU examples in the ``good'' bin and highest scoring BLEU examples in the ``bad'' bin for \texttt{\small zh-en} translation evaluation in WMT'17. Examples are binned by DA scores, and chosen from the top/bottom 10. Some words have been elided. Readers are encouraged to view other examples on our notebook.}
    \label{qual-analysis}
\end{table*}

\subsection{Segment-level analysis: metric score distributions conditional on DA}

To understand how your metric is scoring with respect to translations of different quality, you may visualize the distributions of your metric score on ``bins'' of DA scores.  Start by partitioning your examples by DA scores into a lower one-third ($\frac{1}{3}$) DA scores bin (bad translations), a medium one-third bin (average), and higher one-third bin (good). Each bin should have a near equal number of points. Produce violin plots of the conditional distributions as in \citet{DBLP:conf/acl/LiangCM18}. Examining conditional correlations within bins \emph{does not} provide the same information!

Refer to Figure \ref{bleu-cond-dist}. We see that sentence-BLEU has significant overlap in scoring bad and average translations, but not the good translation. Our interpretation is that sentence-BLEU mainly differentiates good translations well. Points that lay high on the $x$-axis (high $n$-gram overlap) will almost always be good translations, as the reference and system output will be nearly exact matches. In \texttt{*-en} translation, this strategy is effective to achieve high system-level correlation.

\subsection{Segment-level analysis: qualitative analysis on metric failure cases}

One of the best ways to understand your metric is to examine failure cases - cases in which there is large disagreement between human DA scores and your metric score. We have found this analysis to be insightful even looking at 5-10 examples. After binning examples into three bins by DA score (as in \S 4.1), sort the ``good'' bin in ascending order, and the ``bad'' bin in descending order. You may also want to produce some features of your metric for comparison. Qualitative analysis has been used in \citet{DBLP:conf/aaai/TaoMZY18}.

Refer to Table \ref{qual-analysis}. One of the biggest shortcomings of BLEU is the lack of respect to the semantic content. In the good system outputs that received low BLEU scores, we see several meaning preserving paraphrases of the reference. However, we note that these instances are quite rare (observe Figure \ref{bleu-cond-dist}). In the bad output examples, there are at least two failure cases of BLEU: matching some words in really short sentences and repeating long phrases artificially inflates the BLEU score.

\begin{figure}
    \centering
    \includegraphics[scale=0.75]{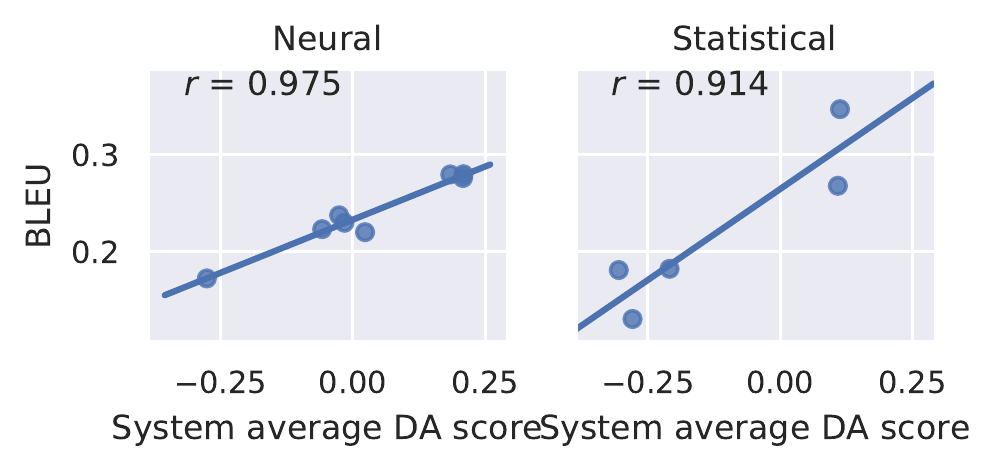}
    \caption{Sample Pearson correlation of BLEU with DA conditional on system type for \texttt{\small zh-en} translation evaluation in WMT'17. System types were classified by hand. Four out of five statistical systems were anonymized online translation systems.  }
    \label{neural-vs-statistical}
\end{figure}

\subsection{System-level analysis: metric correlation with DA conditional on system type}

The correlation of a metric can vary greatly depending on the diversity of systems it is validated on. As shown in \citet{DBLP:conf/eacl/Callison-BurchOK06}, your correlation may decrease when validated against a diverse range of systems. In practice, if we know that a metric has weak correlation for comparing systems with different approaches, we may want to constrain metric use to comparing systems using the same approach (e.g. neural).

Refer to Figure \ref{neural-vs-statistical}. The correlation for neural systems is higher than that of the statistical systems. These recent switch from statistical to neural MT systems is a likely factor in \citet{DBLP:journals/coling/Reiter18} observing human-BLEU correlation increasing over time. When comparing BLEU scores, it is more effective to compare neural systems to other neural systems. In validating our metrics, we must choose a mix of possible approaches to better understand our correlation. 

\begin{figure}[]
    \centering
    \hspace*{-0.35cm}
    \includegraphics[scale=0.75]{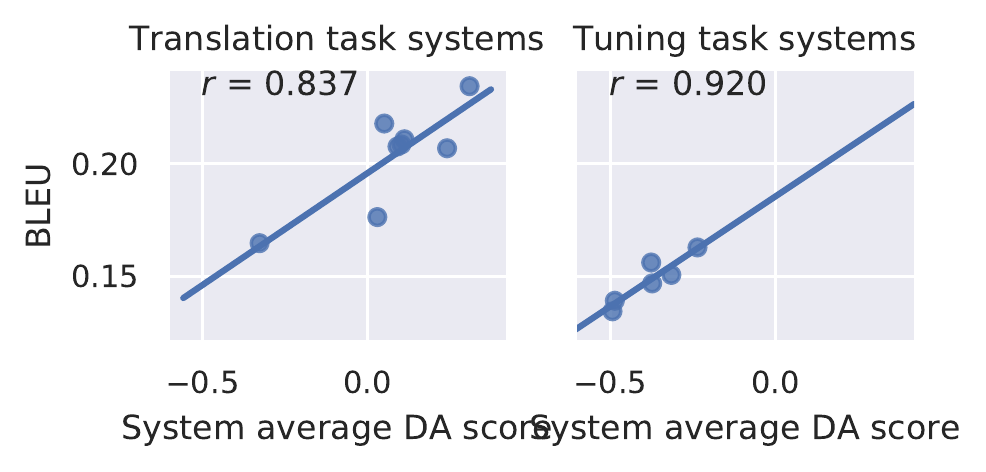}
    \caption{Sample Pearson correlation of BLEU with DA conditional on submission track for \texttt{en-cs} translation evaluation in WMT'17. All tuning task systems are instances of Neural Monkey. \cite{NeuralMonkey:2017} The translation task includes different systems. }
    \label{neuralmonkey-vs-other}
\end{figure}

Refer to Figure \ref{neuralmonkey-vs-other}. The WMT'17 tuning task \cite{DBLP:conf/wmt/BojarHKLM17} calls for participants to tune the same neural MT model \cite[Neural Monkey;][]{NeuralMonkey:2017} for \texttt{\small en-cs} translation with varying training settings e.g. curriculum learning etc. Intuitively, models of the same architecture are likely to make similar mistakes, on average, over the same test set, so are penalized by humans equally. This causes the spread (or residuals) of the best-fit-line to be much tighter. 

\begin{table}[]
    \centering
    \begin{tabular}{c|c|c|c}
        ~ & $n$ & $|r|$ & $|\tau|$ \\
        \hline
        \texttt{\small cs-en} & 4 & 0.971 & 1.000 \\
        \texttt{\small de-en} & 11 & 0.923 & 0.564 \\
        \texttt{\small fi-en} & 6 & 0.903 & 0.867 \\
        \texttt{\small lv-en} & 9 & 0.979 & 0.833 \\
        \texttt{\small ru-en} & 9 & 0.912 & 0.778 \\
        \texttt{\small tr-en} & 10 & 0.976 & 0.911 \\
        \texttt{\small zh-en} & 16 & 0.864 & 0.767 \\
    \end{tabular}
    \caption{Sample Pearson's $r$ and Kendall's $\tau$ correlations between BLEU and DA at the system level. }
    \label{kendalls-tau}
\end{table}

\subsection{System-level analysis: Kendall's $\tau$ coefficient and its interpretation}

In this section we propose the use of Kendall's $\tau$ as a secondary evaluation metric on the system level. While this coefficient does not have an appropriate significance test in our setting, it has an intuitive interpretation. For paired data points $\{(x_1, y_1), ..., (x_n, y_n)\}$ Kendall's $\tau$ is a rank-based correlation coefficient defined as
$$\frac{\text{(Concordant pairs)-(Discordant pairs)}}{\text{(Concordant pairs)+(Discordant pairs)}}$$
where the number of concordant pairs are the number of pairs such that $x_i < x_j \land y_i < y_j$ or $x_i > x_j \land y_i > y_j$ for all $1\leq i \leq j \leq n$. Pairs are discordant when there is disagreement in rank between the two variables. Kendall's $\tau$ is then a difference in the percentage (normalized over total pairs) of concordant and discordant pairs.

Intuitively, $x_i$ and $y_i$ represent the BLEU and DA scores. Refer to Table \ref{kendalls-tau}. Kendall's $\tau$ then tells us the percentage difference that BLEU would agree over disagree with DA pairwise judgment on system quality. For instance in \texttt{zh-en}, BLEU agrees more than disagrees 86 percentage points with DA for pairwise judgments. We leave the interpretation of correlation metrics more relevant in practice as future work.

\section{Relevant work: metrics}

This section attempts to make a selection of influential work that overviews the metrics development literature. Besides the well-known BLEU metric, other metrics primarily using $n$-gram features include the NIST \cite{Doddington:2002:AEM:1289189.1289273} metric for translation, which is similar to BLEU but weighs $n$-grams based on their rarity, and ROUGE \cite{rouge-a-package-for-automatic-evaluation-of-summaries} for summarization. Character-level $n$-gram features have also been proposed to capture subword information. The {\small CHR}F metric \cite{DBLP:conf/wmt/Popovic15} calculate F-scores based on character-level $n$-grams. CDER and BEER \cite{DBLP:conf/wmt/StanojevicS14} also use character features.

Several translation metrics calculate scores based on alignments of extracted surface linguistic features between hypotheses and references. METEOR \cite{denkowski:lavie:meteor-wmt:2014} uses alignments based on exact, stem, synonym, and paraphrase matches between words and phrases.\footnote{\url{http://www.cs.cmu.edu/~alavie/METEOR/}} MEANT \cite{DBLP:conf/wmt/Lo17} evaluates translations by first aligning semantic frames and then aligning role fillers. Both MEANT and METEOR incorporate linguistic resources such as WordNet and semantic role parsers, respectively, and their incorporation, including other character-level and shallow linguistic approaches, have shown benefits in the WMT metrics task. \cite{bojar-etal_Cracker:2016}

More recently, both formal and distributed semantic representations have found success in metrics, even in NLG domains outside of translation. SPICE \cite{DBLP:conf/eccv/AndersonFJG16} is an image captioning metric based on semantic parsing, where scores are based on overlap of semantic propositions in the graph meaning representations of both the hypothesis and reference captions. The RUBER metric \cite{DBLP:conf/aaai/TaoMZY18} for dialogue uses cosine similarities of word embeddings to predict response appropriateness, and a neural model trained with negative sampling to predict relevancy. The first metric in WMT'18 to use sentence embeddings, RUSE \cite{DBLP:conf/wmt/ShimanakaKK18} scores with a trained neural regression model on sentence embeddings and is the highest performing metric for \texttt{*-en} translation. 

Finally, we will highlight MT quality estimation systems (QE, evaluating how good a translation is without the reference) which includes both neural and grammar-based approaches. Recurrent neural networks encoding both the source and MT output have been used for regression on human quality scores. \cite{DBLP:journals/corr/abs-1708-01759} The WMT quality estimation shared task \cite{DBLP:conf/wmt/SpeciaBLAM18} includes examples of many such systems, with a winning submission using neural models.  \cite{DBLP:conf/wmt/WangFLZCSS18} Approaches based on grammatical error correction systems are reference-less and have also been proposed for QE. \cite{DBLP:conf/emnlp/NapolesST16}
 
\section{Conclusion}

The authors believe a metric that applies across multiple NLG domains may be widely adopted. This metric might focus only on semantic similarity (with a reference), and be complemented with another domain-specific measure e.g. style preservation for style transfer, \cite{DBLP:conf/aaai/FuTPZY18} summary length for summarization, or relevancy for dialogue. \cite{DBLP:conf/aaai/TaoMZY18} To compute semantic similarity, a model may use meaning representations and parsing methods \cite{DBLP:conf/acl/KonstasIYCZ17} or continuous sentence/word embeddings, \cite{Peters:2018, DBLP:conf/emnlp/ConneauKSBB17} as these technologies are becoming more widespread and powerful in NLP.

For a few NLG tasks e.g. MT, image captioning, summarization, achieving usable system-level correlation is not impossible. In some cases, it can be argued that achieving usable system-level correlation is not difficult, as merely correctly differentiating good output is sufficient. (\S 4.1) However, segment-level correlation leaves much to be desired. The existence of metrics with high segment-level correlation opens up exciting research directions. Such a metric can also be useful in practice, as it can identify model failure modes, or detect low quality output and fallback on rule-based models as needed.

\bibliography{naaclhlt2019.bib}

\begin{thebibliography}{54}
\expandafter\ifx\csname natexlab\endcsname\relax\def\natexlab#1{#1}\fi

\bibitem[{Anderson et~al.(2016)Anderson, Fernando, Johnson, and
  Gould}]{DBLP:conf/eccv/AndersonFJG16}
Peter Anderson, Basura Fernando, Mark Johnson, and Stephen Gould. 2016.
\newblock \href {https://doi.org/10.1007/978-3-319-46454-1\_24} {{SPICE:}
  semantic propositional image caption evaluation}.
\newblock In \emph{Computer Vision - {ECCV} 2016 - 14th European Conference,
  Amsterdam, The Netherlands, October 11-14, 2016, Proceedings, Part {V}},
  volume 9909 of \emph{Lecture Notes in Computer Science}, pages 382--398.
  Springer.

\bibitem[{Bojar et~al.(2017{\natexlab{a}})Bojar, Buck, Chatterjee, Federmann,
  Graham, Haddow, Huck, Jimeno{-}Yepes, Koehn, and
  Kreutzer}]{DBLP:conf/wmt/2017}
Ondrej Bojar, Christian Buck, Rajen Chatterjee, Christian Federmann, Yvette
  Graham, Barry Haddow, Matthias Huck, Antonio Jimeno{-}Yepes, Philipp Koehn,
  and Julia Kreutzer, editors. 2017{\natexlab{a}}.
\newblock \href
  {http://aclanthology.info/volumes/proceedings-of-the-second-conference-on-machine-translation}
  {\emph{Proceedings of the Second Conference on Machine Translation, {WMT}
  2017, Copenhagen, Denmark, September 7-8, 2017}}. Association for
  Computational Linguistics.

\bibitem[{Bojar et~al.(2018{\natexlab{a}})Bojar, Chatterjee, Federmann, Fishel,
  Graham, Haddow, Huck, Jimeno{-}Yepes, Koehn, Monz, Negri, N{\'{e}}v{\'{e}}ol,
  Neves, Post, Specia, Turchi, and Verspoor}]{DBLP:conf/wmt/2018s}
Ondrej Bojar, Rajen Chatterjee, Christian Federmann, Mark Fishel, Yvette
  Graham, Barry Haddow, Matthias Huck, Antonio Jimeno{-}Yepes, Philipp Koehn,
  Christof Monz, Matteo Negri, Aur{\'{e}}lie N{\'{e}}v{\'{e}}ol, Mariana~L.
  Neves, Matt Post, Lucia Specia, Marco Turchi, and Karin Verspoor, editors.
  2018{\natexlab{a}}.
\newblock \href
  {https://aclanthology.info/volumes/proceedings-of-the-third-conference-on-machine-translation-shared-task-papers}
  {\emph{Proceedings of the Third Conference on Machine Translation: Shared
  Task Papers, {WMT} 2018, Belgium, Brussels, October 31 - November 1, 2018}}.
  Association for Computational Linguistics.

\bibitem[{Bojar et~al.(2017{\natexlab{b}})Bojar, Chatterjee, Federmann, Graham,
  Haddow, Huang, Huck, Koehn, Liu, Logacheva, Monz, Negri, Post, Rubino,
  Specia, and Turchi}]{DBLP:conf/wmt/BojarCFGHHHKLLM17}
Ondrej Bojar, Rajen Chatterjee, Christian Federmann, Yvette Graham, Barry
  Haddow, Shujian Huang, Matthias Huck, Philipp Koehn, Qun Liu, Varvara
  Logacheva, Christof Monz, Matteo Negri, Matt Post, Raphael Rubino, Lucia
  Specia, and Marco Turchi. 2017{\natexlab{b}}.
\newblock \href {https://aclanthology.info/papers/W17-4717/w17-4717} {Findings
  of the 2017 conference on machine translation {(WMT17)}}.
\newblock In  \cite{DBLP:conf/wmt/2017}, pages 169--214.

\bibitem[{Bojar et~al.(2016{\natexlab{a}})Bojar, Chatterjee, Federmann, Graham,
  Haddow, Huck, Jimeno{-}Yepes, Koehn, Logacheva, Monz, Negri,
  N{\'{e}}v{\'{e}}ol, Neves, Popel, Post, Rubino, Scarton, Specia, Turchi,
  Verspoor, and Zampieri}]{DBLP:conf/wmt/BojarCFGHHJKLMN16}
Ondrej Bojar, Rajen Chatterjee, Christian Federmann, Yvette Graham, Barry
  Haddow, Matthias Huck, Antonio Jimeno{-}Yepes, Philipp Koehn, Varvara
  Logacheva, Christof Monz, Matteo Negri, Aur{\'{e}}lie N{\'{e}}v{\'{e}}ol,
  Mariana~L. Neves, Martin Popel, Matt Post, Raphael Rubino, Carolina Scarton,
  Lucia Specia, Marco Turchi, Karin~M. Verspoor, and Marcos Zampieri.
  2016{\natexlab{a}}.
\newblock \href {http://aclweb.org/anthology/W/W16/W16-2301.pdf} {Findings of
  the 2016 conference on machine translation}.
\newblock In \emph{Proceedings of the First Conference on Machine Translation,
  {WMT} 2016, colocated with {ACL} 2016, August 11-12, Berlin, Germany}, pages
  131--198. The Association for Computer Linguistics.

\bibitem[{Bojar et~al.(2018{\natexlab{b}})Bojar, Federmann, Fishel, Graham,
  Haddow, Koehn, and Monz}]{DBLP:conf/wmt/BojarFFGHKM18}
Ondrej Bojar, Christian Federmann, Mark Fishel, Yvette Graham, Barry Haddow,
  Philipp Koehn, and Christof Monz. 2018{\natexlab{b}}.
\newblock \href {https://aclanthology.info/papers/W18-6401/w18-6401} {Findings
  of the 2018 conference on machine translation {(WMT18)}}.
\newblock In  \cite{DBLP:conf/wmt/2018s}, pages 272--303.

\bibitem[{Bojar et~al.(2017{\natexlab{c}})Bojar, Graham, and
  Kamran}]{DBLP:conf/wmt/BojarGK17}
Ondrej Bojar, Yvette Graham, and Amir Kamran. 2017{\natexlab{c}}.
\newblock \href {https://aclanthology.info/papers/W17-4755/w17-4755} {Results
  of the {WMT17} metrics shared task}.
\newblock In  \cite{DBLP:conf/wmt/2017}, pages 489--513.

\bibitem[{Bojar et~al.(2017{\natexlab{d}})Bojar, Helcl, Kocmi, Libovick{\'{y}},
  and Musil}]{DBLP:conf/wmt/BojarHKLM17}
Ondrej Bojar, Jindrich Helcl, Tom Kocmi, Jindrich Libovick{\'{y}}, and
  Tom{\'{a}}s Musil. 2017{\natexlab{d}}.
\newblock \href {https://aclanthology.info/papers/W17-4757/w17-4757} {Results
  of the {WMT17} neural {MT} training task}.
\newblock In  \cite{DBLP:conf/wmt/2017}, pages 525--533.

\bibitem[{Bojar et~al.(2016{\natexlab{b}})Bojar, Federmann, Haddow, Koehn, and
  Matt Post~and}]{bojar-etal_Cracker:2016}
Ond\v{r}ej Bojar, Christian Federmann, Barry Haddow, Philipp Koehn, and
  Lucia~Specia Matt Post~and. 2016{\natexlab{b}}.
\newblock \href
  {http://www.cracking-the-language-barrier.eu/wp-content/uploads/LREC-2016-MT-Eval-Workshop-Proceedings.pdf}
  {Ten years of wmt evaluation campaigns: Lessons learnt}.
\newblock In \emph{Workshop on Translation Evaluation: From Fragmented Tools
  and Data Sets to an Integrated Ecosystem}, pages 27--36, Portoroz, Slovenia.

\bibitem[{Britz et~al.(2017)Britz, Goldie, Luong, and Le}]{D17-1151}
Denny Britz, Anna Goldie, Minh-Thang Luong, and Quoc Le. 2017.
\newblock \href {https://doi.org/10.18653/v1/D17-1151} {Massive exploration of
  neural machine translation architectures}.
\newblock In \emph{Proceedings of the 2017 Conference on Empirical Methods in
  Natural Language Processing}, pages 1442--1451. Association for Computational
  Linguistics.

\bibitem[{Callison{-}Burch et~al.(2006)Callison{-}Burch, Osborne, and
  Koehn}]{DBLP:conf/eacl/Callison-BurchOK06}
Chris Callison{-}Burch, Miles Osborne, and Philipp Koehn. 2006.
\newblock \href {http://aclweb.org/anthology/E/E06/E06-1032.pdf} {Re-evaluation
  the role of bleu in machine translation research}.
\newblock In \emph{{EACL} 2006, 11st Conference of the European Chapter of the
  Association for Computational Linguistics, Proceedings of the Conference,
  April 3-7, 2006, Trento, Italy}. The Association for Computer Linguistics.

\bibitem[{Chaganty et~al.(2018)Chaganty, Mussmann, and
  Liang}]{DBLP:conf/acl/LiangCM18}
Arun~Tejasvi Chaganty, Stephen Mussmann, and Percy Liang. 2018.
\newblock \href {https://aclanthology.info/papers/P18-1060/p18-1060} {The price
  of debiasing automatic metrics in natural language evalaution}.
\newblock In \emph{Proceedings of the 56th Annual Meeting of the Association
  for Computational Linguistics, {ACL} 2018, Melbourne, Australia, July 15-20,
  2018, Volume 1: Long Papers}, pages 643--653. Association for Computational
  Linguistics.

\bibitem[{Conneau et~al.(2017)Conneau, Kiela, Schwenk, Barrault, and
  Bordes}]{DBLP:conf/emnlp/ConneauKSBB17}
Alexis Conneau, Douwe Kiela, Holger Schwenk, Lo{\"{\i}}c Barrault, and Antoine
  Bordes. 2017.
\newblock \href {https://aclanthology.info/papers/D17-1070/d17-1070}
  {Supervised learning of universal sentence representations from natural
  language inference data}.
\newblock In  \cite{DBLP:conf/emnlp/2017}, pages 670--680.

\bibitem[{Denkowski and Lavie(2014)}]{denkowski:lavie:meteor-wmt:2014}
Michael Denkowski and Alon Lavie. 2014.
\newblock Meteor universal: Language specific translation evaluation for any
  target language.
\newblock In \emph{Proceedings of the EACL 2014 Workshop on Statistical Machine
  Translation}.

\bibitem[{Doddington(2002)}]{Doddington:2002:AEM:1289189.1289273}
George Doddington. 2002.
\newblock \href {http://dl.acm.org/citation.cfm?id=1289189.1289273} {Automatic
  evaluation of machine translation quality using n-gram co-occurrence
  statistics}.
\newblock In \emph{Proceedings of the Second International Conference on Human
  Language Technology Research}, HLT '02, pages 138--145, San Francisco, CA,
  USA. Morgan Kaufmann Publishers Inc.

\bibitem[{Dusek et~al.(2017)Dusek, Novikova, and
  Rieser}]{DBLP:journals/corr/abs-1708-01759}
Ondrej Dusek, Jekaterina Novikova, and Verena Rieser. 2017.
\newblock \href {http://arxiv.org/abs/1708.01759} {Referenceless quality
  estimation for natural language generation}.
\newblock \emph{CoRR}, abs/1708.01759.

\bibitem[{Fu et~al.(2018)Fu, Tan, Peng, Zhao, and
  Yan}]{DBLP:conf/aaai/FuTPZY18}
Zhenxin Fu, Xiaoye Tan, Nanyun Peng, Dongyan Zhao, and Rui Yan. 2018.
\newblock \href
  {https://www.aaai.org/ocs/index.php/AAAI/AAAI18/paper/view/17015} {Style
  transfer in text: Exploration and evaluation}.
\newblock In  \cite{DBLP:conf/aaai/2018}, pages 663--670.

\bibitem[{Gatt and Krahmer(2018)}]{DBLP:journals/jair/GattK18}
Albert Gatt and Emiel Krahmer. 2018.
\newblock \href {https://doi.org/10.1613/jair.5477} {Survey of the state of the
  art in natural language generation: Core tasks, applications and evaluation}.
\newblock \emph{J. Artif. Intell. Res.}, 61:65--170.

\bibitem[{Gkatzia and Mahamood(2015)}]{DBLP:conf/enlg/GkatziaM15}
Dimitra Gkatzia and Saad Mahamood. 2015.
\newblock \href {http://aclweb.org/anthology/W/W15/W15-4708.pdf} {A snapshot of
  {NLG} evaluation practices 2005 - 2014}.
\newblock In \emph{{ENLG} 2015 - Proceedings of the 15th European Workshop on
  Natural Language Generation, 10-11 September 2015, University of Brighton,
  Brighton, {UK}}, pages 57--60. The Association for Computer Linguistics.

\bibitem[{Graham et~al.(2018)Graham, Awad, and
  Smeaton}]{10.1371/journal.pone.0202789}
Yvette Graham, George Awad, and Alan Smeaton. 2018.
\newblock \href {https://doi.org/10.1371/journal.pone.0202789} {Evaluation of
  automatic video captioning using direct assessment}.
\newblock \emph{PLOS ONE}, 13(9):1--20.

\bibitem[{Graham and Baldwin(2014)}]{DBLP:conf/emnlp/GrahamB14}
Yvette Graham and Timothy Baldwin. 2014.
\newblock \href {http://aclweb.org/anthology/D/D14/D14-1020.pdf} {Testing for
  significance of increased correlation with human judgment}.
\newblock In \emph{Proceedings of the 2014 Conference on Empirical Methods in
  Natural Language Processing, {EMNLP} 2014, October 25-29, 2014, Doha, Qatar,
  {A} meeting of SIGDAT, a Special Interest Group of the {ACL}}, pages
  172--176. {ACL}.

\bibitem[{Graham et~al.(2015)Graham, Baldwin, and
  Mathur}]{DBLP:conf/naacl/GrahamBM15}
Yvette Graham, Timothy Baldwin, and Nitika Mathur. 2015.
\newblock \href {http://aclweb.org/anthology/N/N15/N15-1124.pdf} {Accurate
  evaluation of segment-level machine translation metrics}.
\newblock In \emph{{NAACL} {HLT} 2015, The 2015 Conference of the North
  American Chapter of the Association for Computational Linguistics: Human
  Language Technologies, Denver, Colorado, USA, May 31 - June 5, 2015}, pages
  1183--1191. The Association for Computational Linguistics.

\bibitem[{Graham et~al.(2013)Graham, Baldwin, Moffat, and
  Zobel}]{DBLP:conf/acllaw/GrahamBMZ13}
Yvette Graham, Timothy Baldwin, Alistair Moffat, and Justin Zobel. 2013.
\newblock \href {http://aclweb.org/anthology/W/W13/W13-2305.pdf} {Continuous
  measurement scales in human evaluation of machine translation}.
\newblock In \emph{Proceedings of the 7th Linguistic Annotation Workshop and
  Interoperability with Discourse, LAW-ID@ACL 2013, August 8-9, 2013, Sofia,
  Bulgaria}, pages 33--41. The Association for Computer Linguistics.

\bibitem[{Helcl and Libovick{\'{y}}(2017)}]{NeuralMonkey:2017}
Jind{\v{r}}ich Helcl and Jind{\v{r}}ich Libovick{\'{y}}. 2017.
\newblock \href {https://doi.org/10.1515/pralin-2017-0001} {{Neural Monkey: An
  Open-source Tool for Sequence Learning}}.
\newblock \emph{The Prague Bulletin of Mathematical Linguistics}, (107):5--17.

\bibitem[{Kilickaya et~al.(2017)Kilickaya, Erdem, Ikizler{-}Cinbis, and
  Erdem}]{DBLP:conf/eacl/KilickayaEIE17}
Mert Kilickaya, Aykut Erdem, Nazli Ikizler{-}Cinbis, and Erkut Erdem. 2017.
\newblock \href {https://aclanthology.info/papers/E17-1019/e17-1019}
  {Re-evaluating automatic metrics for image captioning}.
\newblock In \emph{Proceedings of the 15th Conference of the European Chapter
  of the Association for Computational Linguistics, {EACL} 2017, Valencia,
  Spain, April 3-7, 2017, Volume 1: Long Papers}, pages 199--209. Association
  for Computational Linguistics.

\bibitem[{Koehn(2004)}]{DBLP:conf/emnlp/Koehn04}
Philipp Koehn. 2004.
\newblock \href {http://www.aclweb.org/anthology/W04-3250} {Statistical
  significance tests for machine translation evaluation}.
\newblock In \emph{Proceedings of the 2004 Conference on Empirical Methods in
  Natural Language Processing , {EMNLP} 2004, {A} meeting of SIGDAT, a Special
  Interest Group of the ACL, held in conjunction with {ACL} 2004, 25-26 July
  2004, Barcelona, Spain}, pages 388--395. {ACL}.

\bibitem[{Koehn et~al.(2007)Koehn, Hoang, Birch, Callison-Burch, Federico,
  Bertoldi, Cowan, Shen, Moran, Zens, Dyer, Bojar, Constantin, and
  Herbst}]{Koehn:2007:MOS:1557769.1557821}
Philipp Koehn, Hieu Hoang, Alexandra Birch, Chris Callison-Burch, Marcello
  Federico, Nicola Bertoldi, Brooke Cowan, Wade Shen, Christine Moran, Richard
  Zens, Chris Dyer, Ond\v{r}ej Bojar, Alexandra Constantin, and Evan Herbst.
  2007.
\newblock \href {http://dl.acm.org/citation.cfm?id=1557769.1557821} {Moses:
  Open source toolkit for statistical machine translation}.
\newblock In \emph{Proceedings of the 45th Annual Meeting of the ACL on
  Interactive Poster and Demonstration Sessions}, ACL '07, pages 177--180,
  Stroudsburg, PA, USA. Association for Computational Linguistics.

\bibitem[{Konstas et~al.(2017)Konstas, Iyer, Yatskar, Choi, and
  Zettlemoyer}]{DBLP:conf/acl/KonstasIYCZ17}
Ioannis Konstas, Srinivasan Iyer, Mark Yatskar, Yejin Choi, and Luke
  Zettlemoyer. 2017.
\newblock \href {https://doi.org/10.18653/v1/P17-1014} {Neural {AMR:}
  sequence-to-sequence models for parsing and generation}.
\newblock In \emph{Proceedings of the 55th Annual Meeting of the Association
  for Computational Linguistics, {ACL} 2017, Vancouver, Canada, July 30 -
  August 4, Volume 1: Long Papers}, pages 146--157. Association for
  Computational Linguistics.

\bibitem[{Lin(2004)}]{rouge-a-package-for-automatic-evaluation-of-summaries}
Chin-Yew Lin. 2004.
\newblock \href
  {https://www.microsoft.com/en-us/research/publication/rouge-a-package-for-automatic-evaluation-of-summaries/}
  {Rouge: a package for automatic evaluation of summaries}.

\bibitem[{Lin et~al.(2014)Lin, Maire, Belongie, Hays, Perona, Ramanan,
  Doll{\'{a}}r, and Zitnick}]{DBLP:conf/eccv/LinMBHPRDZ14}
Tsung{-}Yi Lin, Michael Maire, Serge~J. Belongie, James Hays, Pietro Perona,
  Deva Ramanan, Piotr Doll{\'{a}}r, and C.~Lawrence Zitnick. 2014.
\newblock \href {https://doi.org/10.1007/978-3-319-10602-1\_48} {Microsoft
  {COCO:} common objects in context}.
\newblock In \emph{Computer Vision - {ECCV} 2014 - 13th European Conference,
  Zurich, Switzerland, September 6-12, 2014, Proceedings, Part {V}}, volume
  8693 of \emph{Lecture Notes in Computer Science}, pages 740--755. Springer.

\bibitem[{Liu et~al.(2016)Liu, Lowe, Serban, Noseworthy, Charlin, and
  Pineau}]{DBLP:conf/emnlp/LiuLSNCP16}
Chia{-}Wei Liu, Ryan Lowe, Iulian Serban, Michael Noseworthy, Laurent Charlin,
  and Joelle Pineau. 2016.
\newblock \href {http://aclweb.org/anthology/D/D16/D16-1230.pdf} {How {NOT} to
  evaluate your dialogue system: An empirical study of unsupervised evaluation
  metrics for dialogue response generation}.
\newblock In  \cite{DBLP:conf/emnlp/2016}, pages 2122--2132.

\bibitem[{Lo(2017)}]{DBLP:conf/wmt/Lo17}
Chi{-}kiu Lo. 2017.
\newblock \href {https://aclanthology.info/papers/W17-4767/w17-4767} {{MEANT}
  2.0: Accurate semantic {MT} evaluation for any output language}.
\newblock In  \cite{DBLP:conf/wmt/2017}, pages 589--597.

\bibitem[{Ma et~al.(2018)Ma, Bojar, and Graham}]{DBLP:conf/wmt/MaBG18}
Qingsong Ma, Ondrej Bojar, and Yvette Graham. 2018.
\newblock \href {https://aclanthology.info/papers/W18-6450/w18-6450} {Results
  of the {WMT18} metrics shared task: Both characters and embeddings achieve
  good performance}.
\newblock In  \cite{DBLP:conf/wmt/2018s}, pages 671--688.

\bibitem[{McIlraith and Weinberger(2018)}]{DBLP:conf/aaai/2018}
Sheila~A. McIlraith and Kilian~Q. Weinberger, editors. 2018.
\newblock \href
  {https://www.aaai.org/ocs/index.php/AAAI/AAAI18/schedConf/presentations}
  {\emph{Proceedings of the Thirty-Second {AAAI} Conference on Artificial
  Intelligence, (AAAI-18), the 30th innovative Applications of Artificial
  Intelligence (IAAI-18), and the 8th {AAAI} Symposium on Educational Advances
  in Artificial Intelligence (EAAI-18), New Orleans, Louisiana, USA, February
  2-7, 2018}}. {AAAI} Press.

\bibitem[{Napoles et~al.(2016)Napoles, Sakaguchi, and
  Tetreault}]{DBLP:conf/emnlp/NapolesST16}
Courtney Napoles, Keisuke Sakaguchi, and Joel~R. Tetreault. 2016.
\newblock \href {http://aclweb.org/anthology/D/D16/D16-1228.pdf} {There's no
  comparison: Reference-less evaluation metrics in grammatical error
  correction}.
\newblock In  \cite{DBLP:conf/emnlp/2016}, pages 2109--2115.

\bibitem[{Novikova et~al.(2017)Novikova, Dusek, Curry, and
  Rieser}]{DBLP:conf/emnlp/NovikovaDCR17}
Jekaterina Novikova, Ondrej Dusek, Amanda~Cercas Curry, and Verena Rieser.
  2017.
\newblock \href {https://aclanthology.info/papers/D17-1238/d17-1238} {Why we
  need new evaluation metrics for {NLG}}.
\newblock In  \cite{DBLP:conf/emnlp/2017}, pages 2241--2252.

\bibitem[{Palmer et~al.(2017)Palmer, Hwa, and Riedel}]{DBLP:conf/emnlp/2017}
Martha Palmer, Rebecca Hwa, and Sebastian Riedel, editors. 2017.
\newblock \href
  {https://aclanthology.info/volumes/proceedings-of-the-2017-conference-on-empirical-methods-in-natural-language-processing}
  {\emph{Proceedings of the 2017 Conference on Empirical Methods in Natural
  Language Processing, {EMNLP} 2017, Copenhagen, Denmark, September 9-11,
  2017}}. Association for Computational Linguistics.

\bibitem[{Papineni et~al.(2002)Papineni, Roukos, Ward, and
  Zhu}]{DBLP:conf/acl/PapineniRWZ02}
Kishore Papineni, Salim Roukos, Todd Ward, and Wei{-}Jing Zhu. 2002.
\newblock \href {http://www.aclweb.org/anthology/P02-1040.pdf} {Bleu: a method
  for automatic evaluation of machine translation}.
\newblock In \emph{Proceedings of the 40th Annual Meeting of the Association
  for Computational Linguistics, July 6-12, 2002, Philadelphia, PA, {USA.}},
  pages 311--318. {ACL}.

\bibitem[{Peters et~al.(2018)Peters, Neumann, Iyyer, Gardner, Clark, Lee, and
  Zettlemoyer}]{Peters:2018}
Matthew~E. Peters, Mark Neumann, Mohit Iyyer, Matt Gardner, Christopher Clark,
  Kenton Lee, and Luke Zettlemoyer. 2018.
\newblock Deep contextualized word representations.
\newblock In \emph{Proc. of NAACL}.

\bibitem[{Popovic(2015)}]{DBLP:conf/wmt/Popovic15}
Maja Popovic. 2015.
\newblock \href {http://aclweb.org/anthology/W/W15/W15-3049.pdf} {chrf:
  character n-gram f-score for automatic {MT} evaluation}.
\newblock In \emph{Proceedings of the Tenth Workshop on Statistical Machine
  Translation, WMT@EMNLP 2015, 17-18 September 2015, Lisbon, Portugal}, pages
  392--395. The Association for Computer Linguistics.

\bibitem[{Post(2018)}]{DBLP:conf/wmt/Post18}
Matt Post. 2018.
\newblock \href {https://aclanthology.info/papers/W18-6319/w18-6319} {A call
  for clarity in reporting {BLEU} scores}.
\newblock In \emph{Proceedings of the Third Conference on Machine Translation:
  Research Papers, {WMT} 2018, Belgium, Brussels, October 31 - November 1,
  2018}, pages 186--191. Association for Computational Linguistics.

\bibitem[{Ranzato et~al.(2015)Ranzato, Chopra, Auli, and
  Zaremba}]{DBLP:journals/corr/RanzatoCAZ15}
Marc'Aurelio Ranzato, Sumit Chopra, Michael Auli, and Wojciech Zaremba. 2015.
\newblock \href {http://arxiv.org/abs/1511.06732} {Sequence level training with
  recurrent neural networks}.
\newblock \emph{CoRR}, abs/1511.06732.

\bibitem[{Reiter(2018)}]{DBLP:journals/coling/Reiter18}
Ehud Reiter. 2018.
\newblock \href {https://doi.org/10.1162/coli\_a\_00322} {A structured review
  of the validity of {BLEU}}.
\newblock \emph{Computational Linguistics}, 44(3).

\bibitem[{Reiter et~al.(2003)Reiter, Robertson, and
  Osman}]{DBLP:journals/ai/ReiterRO03}
Ehud Reiter, Roma Robertson, and Liesl Osman. 2003.
\newblock \href {https://doi.org/10.1016/S0004-3702(02)00370-3} {Lessons from a
  failure: Generating tailored smoking cessation letters}.
\newblock \emph{Artif. Intell.}, 144(1-2):41--58.

\bibitem[{Sennrich et~al.(2017)Sennrich, Birch, Currey, Germann, Haddow,
  Heafield, Barone, and Williams}]{DBLP:conf/wmt/SennrichBCGHHBW17}
Rico Sennrich, Alexandra Birch, Anna Currey, Ulrich Germann, Barry Haddow,
  Kenneth Heafield, Antonio Valerio~Miceli Barone, and Philip Williams. 2017.
\newblock \href {https://aclanthology.info/papers/W17-4739/w17-4739} {The
  university of edinburgh's neural {MT} systems for {WMT17}}.
\newblock In  \cite{DBLP:conf/wmt/2017}, pages 389--399.

\bibitem[{Shimanaka et~al.(2018)Shimanaka, Kajiwara, and
  Komachi}]{DBLP:conf/wmt/ShimanakaKK18}
Hiroki Shimanaka, Tomoyuki Kajiwara, and Mamoru Komachi. 2018.
\newblock \href {https://aclanthology.info/papers/W18-6456/w18-6456} {{RUSE:}
  regressor using sentence embeddings for automatic machine translation
  evaluation}.
\newblock In  \cite{DBLP:conf/wmt/2018s}, pages 751--758.

\bibitem[{Specia et~al.(2018)Specia, Blain, Logacheva, Astudillo, and
  Martins}]{DBLP:conf/wmt/SpeciaBLAM18}
Lucia Specia, Fr{\'{e}}d{\'{e}}ric Blain, Varvara Logacheva,
  Ram{\'{o}}n~Fern{\'{a}}ndez Astudillo, and Andr{\'{e}} F.~T. Martins. 2018.
\newblock \href {https://aclanthology.info/papers/W18-6451/w18-6451} {Findings
  of the {WMT} 2018 shared task on quality estimation}.
\newblock In  \cite{DBLP:conf/wmt/2018s}, pages 689--709.

\bibitem[{Stanojevic and Sima'an(2014)}]{DBLP:conf/wmt/StanojevicS14}
Milos Stanojevic and Khalil Sima'an. 2014.
\newblock \href {http://aclweb.org/anthology/W/W14/W14-3354.pdf} {{BEER:}
  better evaluation as ranking}.
\newblock In \emph{Proceedings of the Ninth Workshop on Statistical Machine
  Translation, WMT@ACL 2014, June 26-27, 2014, Baltimore, Maryland, {USA}},
  pages 414--419. The Association for Computer Linguistics.

\bibitem[{Su et~al.(2016)Su, Carreras, and Duh}]{DBLP:conf/emnlp/2016}
Jian Su, Xavier Carreras, and Kevin Duh, editors. 2016.
\newblock \href {http://aclweb.org/anthology/D/D16/} {\emph{Proceedings of the
  2016 Conference on Empirical Methods in Natural Language Processing, {EMNLP}
  2016, Austin, Texas, USA, November 1-4, 2016}}. The Association for
  Computational Linguistics.

\bibitem[{Tao et~al.(2018)Tao, Mou, Zhao, and Yan}]{DBLP:conf/aaai/TaoMZY18}
Chongyang Tao, Lili Mou, Dongyan Zhao, and Rui Yan. 2018.
\newblock \href
  {https://www.aaai.org/ocs/index.php/AAAI/AAAI18/paper/view/16179} {{RUBER:}
  an unsupervised method for automatic evaluation of open-domain dialog
  systems}.
\newblock In  \cite{DBLP:conf/aaai/2018}, pages 722--729.

\bibitem[{Vinyals and Le(2015)}]{DBLP:journals/corr/VinyalsL15}
Oriol Vinyals and Quoc~V. Le. 2015.
\newblock \href {http://arxiv.org/abs/1506.05869} {A neural conversational
  model}.
\newblock \emph{CoRR}, abs/1506.05869.

\bibitem[{Wang et~al.(2018)Wang, Fan, Li, Zhou, Chen, Shi, and
  Si}]{DBLP:conf/wmt/WangFLZCSS18}
Jiayi Wang, Kai Fan, Bo~Li, Fengming Zhou, Boxing Chen, Yangbin Shi, and Luo
  Si. 2018.
\newblock \href {https://aclanthology.info/papers/W18-6465/w18-6465} {Alibaba
  submission for {WMT18} quality estimation task}.
\newblock In  \cite{DBLP:conf/wmt/2018s}, pages 809--815.

\bibitem[{Williams(1959)}]{williams1959regression}
E.J. Williams. 1959.
\newblock \href {https://books.google.com.fj/books?id=HXdqAAAAMAAJ}
  {\emph{Regression analysis}}.
\newblock Wiley series in probability and mathematical statistics. Probability
  and mathematical statistics. Wiley.

\bibitem[{Yao et~al.(2018)Yao, Peng, Weischedel, Knight, Zhao, and
  Yan}]{DBLP:journals/corr/abs-1811-05701}
Lili Yao, Nanyun Peng, Ralph~M. Weischedel, Kevin Knight, Dongyan Zhao, and Rui
  Yan. 2018.
\newblock \href {http://arxiv.org/abs/1811.05701} {Plan-and-write: Towards
  better automatic storytelling}.
\newblock \emph{CoRR}, abs/1811.05701.

\end{thebibliography}
\bibliographystyle{acl_natbib}

\end{document}